\documentclass{article}

\usepackage[english]{babel}
\usepackage{todonotes}
\usepackage{tikz}
\usetikzlibrary{patterns}
\usepackage{authblk}
\usepackage{pgfplots}
 \usepackage{fullpage}
\usepackage{subcaption}
\usepackage{amsthm}
\usepackage{thmtools}
\usepackage{amsmath,amssymb}
\usepackage{mathtools}
\usepackage{graphicx}
\usepackage[colorlinks=true, allcolors=blue]{hyperref}
\usepackage[noabbrev,nameinlink]{cleveref}

\usepackage{algorithm}
\usepackage{algorithmicx}
\usepackage{algpseudocode}
\usepackage{comment}

\DeclareMathOperator*{\argmin}{arg\,min}
\newtheorem{definition}{Definition}[section]

\newtheorem{conjecture}[definition]{Conjecture}

\title{Optimizing Kernel Discrepancies via Subset Selection}
\author[1]{Deyao Chen \thanks{dc262@st-andrews.ac.uk}}
\author[2]{Fran\c{c}ois Cl\'ement \thanks{francois.clement@uw.edu}}
\author[3]{Carola Doerr \thanks{carola.doerr@lip6.fr}}
\author[4]{Nathan Kirk 
\thanks{nkirk@illinoistech.edu}}
   \affil[1]{Department of Mathematics and Statistics, University of St Andrews, St Andrews, United Kingdom}
  \affil[2]{Department of Mathematics, University of Washington, Seattle, WA 98195, USA}
  \affil[3]{Sorbonne Universit\'e, CNRS, LIP6, Paris, France}
  \affil[4]{Department of Applied Mathematics, Illinois Institute of Technology, Chicago, IL 60616, USA}
\date{\today}

\begin{document}

\maketitle

\begin{abstract}
Kernel discrepancies are a powerful tool for analyzing worst-case errors in quasi-Monte Carlo (QMC)
methods. Building on recent advances in optimizing such discrepancy measures, we extend the subset
selection problem to the setting of kernel discrepancies, selecting an m-element subset from a large
population of size $n \gg m$. We introduce a novel subset selection algorithm applicable to general kernel
discrepancies to efficiently generate low-discrepancy samples from both the uniform distribution on the
unit hypercube, the traditional setting of classical QMC, and from more general distributions $F$ with
known density functions by employing the kernel Stein discrepancy. We also explore the relationship
between the classical $L_2$ star discrepancy and its $L_\infty$ counterpart.
\end{abstract}

\section{Introduction}

\subsection{Background}

Discrepancy notions quantify how well a discrete point set approximates an underlying target probability distribution. Over the past seventy years, numerous such measures have been proposed and have been studied extensively \cite{Pano,KUIPNIED1974,Mat,Nie92} mostly concerning applications to the error in numerically approximating an integral of a function in quasi-Monte Carlo (QMC) methods via the Koksma-Hlawka inequality \cite{Hlawka,Koksma}. They have been shown to be beneficial in a much wider range of applications; for example, in computer vision~\cite{MatBuilder}, financial mathematics~\cite{GalFin}, numerical integration \cite{DickP10}, experimental design~\cite{FangHick2014,SantnerDoE}, optimization~\cite{CauwetCDLRRTTU20,DiedOpti} and motion planning in robotics \cite{chahine2025improvingefficiencysamplingbasedmotion, vanhuffel2025neurallowdiscrepancysequences} to name a few. The classical setting for QMC is the uniform distribution on $[0,1)^d$ and, from a geometric perspective, discrepancy quantifies the error in estimating the volume of subsets of $[0,1)^d$, often convex or axis-aligned, by the proportion of points they contain. Among them, the most prominent is the $L_{\infty}$ star discrepancy, which captures the worst-case absolute difference between the empirical measure of axis-aligned boxes anchored at the origin and the true volume of this box. Formally, the $L_{\infty}$ star discrepancy of a point set $P$ is given by
\begin{equation}\label{def:discr}
d^{*}_{\infty}(P):=\sup_{q \in [0,1)^d}\bigg| \frac{|P\cap [0,q)|}{|P|}-\lambda([0,q))\bigg|    
\end{equation}
where $\lambda$ denotes the Lebesgue measure. Point sets which produce a small discrepancy ``uniformly'' cover the domain and are known as low-discrepancy (LD).
Traditional constructions of LD point sets, such as those by van der Corput~\cite{vdC}, Halton~\cite{Halton60}, and Sobol'~\cite{Sobol}, have $L_\infty$ star discrepancies which achieve near-optimal asymptotic rates of $O(\log(n)^{d-1}/n)$, though the best currently known lower bound remains $\Omega(\log^{d/2+c(d)}(n)/n)$~\cite{BilykSmall}, leaving open whether these constructions are asymptotically optimal. Despite their theoretical appeal, $L_{\infty}$ discrepancy measures are notoriously difficult to compute in practice, with exact computation being NP-hard~\cite{complexity} and even W[1]-hard in the dimension~\cite{W1hard}. The best-known exact algorithm~\cite{DEM} has complexity $O(n^{1 + d/2})$, rendering it impractical beyond moderate dimensions. As a result, heuristic methods such as threshold accepting~\cite{GnewuchWW12} are often used to obtain lower bounds on the discrepancy of a given point set. By a similar argument, directly minimizing the $L_\infty$ star discrepancy poses several obstacles; not only is the computation NP-hard, but the expression is also non-differentiable and thus not compatible with gradient-based optimization methods. Several recent works have proposed \emph{subset selection} as a powerful non-gradient-based strategy for direct optimization of the star discrepancy, with heuristics developed to approximate optimal subsets from larger candidate sets \cite{CDP22,CDP23}. 

More formally, the subset selection problem that we consider in this text is as follows: 

\begin{center}
  \begin{minipage}{0.6\linewidth}
    \itshape
    Given a point set $P \subset \mathbb{R}^d$ of size $n$, the goal is to select a subset $P_m\subset P$ of $m \ll n$ points that minimizes the discrepancy. That is, we wish to solve the following optimization problem:
\begin{equation}\label{eq:optim_problem}
\argmin_{P_m \subseteq P,\ |P_m| = m} d(P_m)
\end{equation}
where $d(P_m)$ is some discrepancy measure and $m$ the fixed target size.
  \end{minipage}
\end{center}

To address the intractability of directly minimizing the \( L_\infty \) star discrepancy, a popular alternative discrepancy measure is the $L_2$ star discrepancy \cite{KIRKPAUSINGERCLEMENT2024, KIRKPAUSINGER2023,Kritz,MatL2}, which replaces the maximum over local discrepancies with an averaged squared integral. The \textit{\( L_2 \) star discrepancy} of a point set \( P = (x^{(i)})_{i\in\{1,2,\ldots,n\}} \subset [0,1)^d \) is defined as
\[
d^*_2(P) \coloneqq \left(\int_{[0,1)^d}\left|\frac{|P\cap [0,q)|}{|P|}-\lambda([0,q))\right|^2\,dq \right)^{1/2},
\]
and can be evaluated in closed form in $O(dn^2)$ time using the Warnock formula~\cite{Warnock}:
\begin{equation}\label{eq:warnock}
(d^*_{2}(P))^2 := \frac{1}{3^d} - \frac{2}{n}\sum_{i=1}^{n}\prod_{k=1}^{d}\frac{1-(x^{(i)}_k)^2}{2} + \frac{1}{n^2}\sum_{i,j=1}^n\prod_{k=1}^d\left(1-\max\left(x^{(i)}_k,x^{(j)}_k\right)\right).
\end{equation}
Importantly, while the \( L_2 \) discrepancy lacks theoretical worst-case guarantees as a proxy for \( L_\infty \), recent empirical results in small dimensions~\cite{Kritz,RUSCHKIRK2024} indicate that minimizing the \( L_2 \) discrepancy often leads to excellent performance under the \( L_\infty \) metric as well. The reasonable approximation of \( L_\infty \) behavior and the improved computational speed are key motivators to extend the subset selection framework to the \( L_2 \) discrepancy setting. 

Discrepancy theory can also be understood through the lens of reproducing kernel Hilbert spaces (RKHS), as first laid out in \cite{hickernell1998generalized}. From this perspective, the $L_2$ star discrepancy is just one example of a broader class of so-called \emph{kernel discrepancies} which are explored further in Section \ref{sec:kerneldisc}. Several other $L_2$ discrepancy measures such as the periodic, unanchored, and symmetric variants fit into this same framework targeting the uniform distribution on $[0,1]^d$, differing only in the choice of kernel; see \cite{clément2025optimizationdiscrepancymeasures} for a comprehensive treatment of the uniform $L_2$ discrepancy variants. This RHKS-based framework is flexible and extends well beyond the uniform distribution. A particularly important discrepancy is derived by designing the kernel appropriately via Stein’s identity \cite{STEIN1972}; namely, the kernel Stein discrepancy (KSD) \cite{GradientFreeKSD2023,MINSteinDisc2019,STEINPOINTS,GorhamMackey2015,LIULEEJORDAN2016,SVGD2016}. The KSD is nowadays a popular statistical tool for measuring how well a point set approximates a general target distribution $F$ and therefore generalizes the idea of classical uniform discrepancy to non-uniform distributions.

\subsection{Our Contribution}

In this paper, we introduce a generalized subset selection method to solve the optimization problem \eqref{eq:optim_problem} for general distributions \( F \) with known density, by optimizing a suitable, computationally tractable \emph{kernel discrepancy} objective function. We focus here on heuristic methods and an overview of possible exact methods which serves as future research is presented in Section \ref{sec:discussion}.

While presenting our method in full generality, we focus on two specific settings. The first, classical in QMC methods, is the unit hypercube with respect to the uniform distribution. In this case, we optimize the \( L_2 \) star discrepancy \eqref{eq:warnock}, but note that any uniform $L_2$ discrepancy could be applied in this setting \cite{clément2025optimizationdiscrepancymeasures}. Using the traditional $L_2$ discrepancy in low dimensions and a weighted version in higher dimensions, we show that we are able to obtain point sets rivaling the best currently known constructions from~\cite{CDP22,RUSCHKIRK2024} \emph{for the $L_{\infty}$ star discrepancy}. These results provide further evidence for the claim that $L_2$ discrepancies are reasonable surrogates for the $L_{\infty}$  discrepancy; or at the very least, better than previously expected. The second setting leverages the KSD which has garnered significant attention in recent years. As another prevalent example of a kernel discrepancy, we employ the KSD as the objective function in our subset selection algorithm to construct low-discrepancy sets for arbitrary target distributions \( F \) with known density and compare against an established benchmark method of Stein Points \cite{STEINPOINTS} which minimizes the KSD in a greedy manner.

\section{Kernel Discrepancies}\label{sec:kerneldisc}

\subsection{A Reproducing Kernel Hilbert Space Approach}

As mentioned in the Introduction, rather than viewing discrepancy theory through geometric notions, one can use kernel methods and a reproducing kernel Hilbert space (RKHS) approach to form a related but distinct family of measures which we refer to as kernel discrepancies~\cite{hickernell1998generalized}. 
Assuming that a function $f$ of interest belongs to a RKHS $\mathcal{H}$ of functions from \( \mathbb{R}^d \rightarrow \mathbb{R} \) equipped with an inner product \(\langle \cdot, \cdot \rangle_{\mathcal{H}}\) and corresponding norm \(\| \cdot \|_{\mathcal{H}}\), one can bound the error of the sample mean approximation of the integral of $f$ by using the Cauchy-Schwarz inequality within \(\mathcal{H}\): 
\begin{equation}\label{eq:problem}
\left| \frac{1}{n} \sum_{i=1}^n f(x^{(i)}) - \int f(x) \, dF(x) \right| \leq \| f \|_{\mathcal{H}} \, d^{\mathcal{H}}_{2,F} \left( P \right).
\end{equation}
In the above, \( d^{\mathcal{H}}_{2,F} \left( P \right) \) denotes the \textit{kernel discrepancy} of sampling nodes $P=(x^{(i)})_{i\in\{1,\ldots, n\}}$, assessing the quality of the discretization of $F$ by nodes $P$. If $k:\mathbb{R}^d\times\mathbb{R}^d\to\mathbb{R}$ is the kernel associated with the RKHS $\mathcal{H}$ and the integrals \( k_F := \int k(x, \cdot) \, dF(x) \in \mathcal{H} \) and \( k_{F, F} := \int k_F \, dF \) are explicitly available, the discrepancy can be calculated directly by
\begin{equation}\label{eq:general_disc}
d^\mathcal{H}_{2, F} \left( P \right) \coloneqq \sqrt{k_{F, F} - \frac{2}{n} \sum_{i=1}^n k_F\left(x^{(i)}\right) + \frac{1}{n^2} \sum_{i,j=1}^n k\left(x^{(i)}, x^{(j)}\right)}.
\end{equation}
Although we present results in this work primarily on the star discrepancy, we note that any uniform $L_2$ discrepancy, e.g., extreme, centered, periodic etc., can be applied in its place by simply changing the kernel function $k$, as summarized in \cite{clément2025optimizationdiscrepancymeasures}.

\subsection{Stein Discrepancy}

The work of~\cite{GorhamMackey2015} introduced a novel family of sample quality measures known as \emph{Stein discrepancies}, which assess the error in approximating the target distribution in \eqref{eq:problem}, notably, without requiring explicit integration under $F$ with density. These discrepancies are grounded in Stein's identity \cite{STEIN1972}, a fundamental result in probability theory that relates the expectation of a function's derivative to properties of the distribution it is sampled from. Several practical variants of Stein discrepancies have been proposed, however, the kernel Stein discrepancy (KSD) has received the most attention due to its closed-form expression involving pairwise kernel evaluations.

More formally, for a Stein operator \( T_F \), Stein's identity ensures that
\[\int T_F[p(x)] \, dF(x) = 0\] for all $p \in \mathcal{F}$
where \( \mathcal{F} \) is a suitable class of smooth functions. This identity enables the design of operators \( T_F \) that characterize the discrepancy between a sample distribution and the target \( F \). When \( \mathcal{F} \) is chosen to be a RHKS \( \mathcal{H} \) with kernel \( k \), the image of \( \mathcal{H} \) under \( T_F \) forms a new RKHS \( \mathcal{H}_0 := T_F [\mathcal{H}] \), with associated kernel $k_0(x, x') = T_F T_F^* [k(x, x')]$,
where \( T_F^* \) is the adjoint operator of \( T_F \), acting on the second argument of the kernel. A commonly used choice is the Langevin Stein operator, defined by $T_F [p(x)] = \nabla \cdot (q(x) p(x)) / q(x)$
where \( \nabla \cdot \) denotes the divergence operator, \( p \) is a vector-valued function in \( \mathcal{H}^d \) and $q$ is the density function of target distribution $F$. This choice leads to the Stein reproducing kernel:
\[
\begin{aligned}
k_0(x, x') =\ & \nabla_x \cdot \nabla_{x'} k(x, x') + \nabla_x k(x, x') \cdot \nabla_{x'} \log q(x') \\
&+ \nabla_{x'} k(x, x') \cdot \nabla_x \log q(x) + k(x, x') \nabla_x \log q(x) \cdot \nabla_{x'} \log q(x').
\end{aligned}
\]
This kernel satisfies two useful properties: \( k_{0,F} := \int k_0(x, \cdot) \, dF(x) = 0 \), and \( k_{0,F,F} := \int k_{0,F}(x) \, dF(x) = 0 \). Consequently, for a given base kernel \( k \), the KSD is computed as
\begin{equation}\label{eq:KSD}
d^{\mathcal{H}_0}_{2,F}(P) := \sqrt{\frac{1}{n^2} \sum_{i, j=1}^n k_0(x^{(i)}, x^{(j)})}.
\end{equation}

\section{Subset Selection via Kernel Discrepancies}

There exist several advantages of using kernel discrepancies over $L_\infty$-based measures in solving the optimization problem~\eqref{eq:optim_problem}; firstly, the computational effort for a single discrepancy evaluation reduces from $\mathcal{O}(n^{1+d/2})$ to $\mathcal{O}(n^2)$, and secondly, kernel discrepancies admit a decomposition that expresses the individual contribution of each point to the total discrepancy. For example, the contribution of a point \( x^{(i)} \in P \) to the $L_2$ star discrepancy~\eqref{eq:warnock} can be written as
\[
w(x^{(i)},P):= \prod_{k=1}^d \left(1-(x_k^{(i)})^2\right) + \prod_{k=1}^d \left(1-x_k^{(i)}\right) + \frac{2}{n} \sum_{\substack{j=1\\j\neq i}}^{n} \prod_{k=1}^{d} \left(1 - \max\left(x^{(i)}_k,x^{(j)}_k\right)\right),
\]
a decomposition recently exploited in a greedy algorithm by Kritzinger~\cite{Kritz}, and later extended ín~\cite{CKritz}.

We now proceed to describe our heuristic method facilitating the generation of high-quality subsets. Our approach is inspired by the swap-based heuristic of \cite{CDP23}, with one key distinction: as per the observation above, by using kernel discrepancies, we can explicitly identify the \emph{best} greedy improvement at each step, thus removing the need to carefully consider swap options.

\subsection{Algorithm Description}\label{sec:algo}

In general, we can reformulate any (squared) kernel discrepancy \eqref{eq:general_disc} as
\[
\left(d^{\mathcal{H}}_{2,F}(P)\right)^2 = c + \sum_{i,j=1}^n V(i, j),
\]
where $c = k_{F,F}$ and
\begin{equation}\label{eq:contribution}
    V(i, j) = 
    \begin{cases} 
        \frac{1}{n^2} k(x^{(i)}, x^{(j)}) & \text{if } i \ne j, \\
        -\frac{2}{n} k_F(x^{(i)}) + \frac{1}{n^2} k(x^{(i)}, x^{(j)}) & \text{if } i = j
    \end{cases}
\end{equation}
where, as above, \( k_F := \int k(x, \cdot) \, dF(x) \in \mathcal{H} \) and \( k_{F, F} := \int k_F \, dF \).

\begin{algorithm}[t]
\caption{Subset Selection via Kernel Discrepancies}
\label{algo}
\begin{algorithmic}[1]
\Require Population $P = (x^{(i)})_{i\in\{1,2,\ldots,n\}}$, subset size $m$, number of global restarts $R_g$, number of local restarts $R_\ell$, number of initial populations $L$
\Ensure Optimized subset $P_{\text{best}} \subset P$, with $|P_{\text{best}}| = m$

\State Initialize $P_{\text{best}} \gets \emptyset$, $d_{\text{best}} \gets 1$

\For{$r = 1$ to $R_g$}
    \For{$i = 1$ to $L$}
        \State Select initial i.i.d random subset $P_{\text{temp},i} \subset P$ of size $m$
    \EndFor
    \State Let $P_{\text{temp}}$ be the set with the smallest $d^*(P_{\text{temp},i})$ for all $i$.
    \State Select initial i.i.d random subset $P_{\text{temp}} \subset P$ of size $m$.
    \For{$\ell = 1$ to $R_\ell$}
        \State Compute $d(P_{\text{temp}})$ and contributions $B_i$ for all $x^{(i)} \in P$
        \Repeat
            \State Find best swap pair $(x^{(h)}, x^{(k)})$ with $x^{(h)} \in P_{\text{temp}}$, $x^{(k)} \notin P_{\text{temp}}$
            \State Compute $d_{\text{new}} \gets d(P_{\text{temp}}) + B_k - B_h + c(h,k)$
            \If{$d_{\text{new}} < d(P_{\text{temp}})$}
                \State $P_{\text{temp}} \gets (P_{\text{temp}} \setminus \{x^{(h)}\}) \cup \{x^{(k)}\}$
            \Else
                \State No improvement found
            \EndIf
        \Until{no improving swap exists}
        \State Randomly perturb $P_{\text{temp}}$ by swapping a small number of points
    \EndFor
    \If{$d(P_{\text{temp}}) < d_{\text{best}}$}
        \State $P_{\text{best}} \gets P_{\text{temp}}$, $d_{\text{best}} \gets d^*(P_{\text{temp}})$
    \EndIf
\EndFor
\State \Return $P_{\text{best}}$
\end{algorithmic}\label{algo:subset}
\end{algorithm}

\paragraph{Swap-Based Heuristic} The main idea is to improve the current $m$-element subset $P_{\text{temp}} \subset P$ by identifying the pair $(x^{(h)}, x^{(k)})$, with $x^{(h)} \in P_{\text{temp}}$ and $x^{(k)} \in P \setminus P_{\text{temp}}$, that solves
\[
\arg\min_{x^{(h)} \in P_{\text{temp}},\, x^{(k)} \in P \setminus P_{\text{temp}}} d\left((P_{\text{temp}} \setminus \{x^{(h)}\}) \cup \{x^{(k)}\}\right)
\]
where $x^{(h)}$ is then replaced by $x^{(k)}$ if this new discrepancy is lower than that of $P_{\text{temp}}$.
These swaps are repeated until no further improvement can be found, at which point the current solution is perturbed in an attempt to escape any local optima (see below). The complete heuristic is summarized in Algorithm \ref{algo:subset}.

Per pair $(x^{(i)}, x^{(j)})$, the cost to evaluate $V(i,j)$ is $\Theta(d)$ 
provided we use (i) product-form kernels as in classical $L_2$ discrepancies, 
or (ii) common KSD kernels with closed-form mixed derivatives 
(e.g., RBF, IMQ). Consequently, computing the full discrepancy $d(P_{\text{temp}})$ has a cost of 
$\Theta(m^2 d)$, while exhaustively checking all possible swaps requires 
$\Theta(m^3 (n - m)d)$ time.
This becomes infeasible even for moderate $m$, as a large number of swaps 
may be needed to reach a local optimum. To mitigate this cost, we maintain an auxiliary array 
$B = (B_i)_{1 \le i \le n}$ that stores the contribution of each point to the 
discrepancy. 
For points currently in the subset $P_{\text{temp}}$, $B_i$ measures their present 
contribution; for points outside the subset, it represents the contribution they 
would make if added without altering the other points. 
These values satisfy
\[
\begin{aligned}
d\big(P_{\text{temp}} \setminus \{x^{(i)}\}\big)
&= d(P_{\text{temp}}) - B_i, && \text{for } x^{(i)} \in P_{\text{temp}},\\
d\big(P_{\text{temp}} \cup \{x^{(i)}\}\big)
&= d(P_{\text{temp}}) + B_i, && \text{for } x^{(i)} \notin P_{\text{temp}}.
\end{aligned}
\]

Each $B_i$ can be computed by summing over all pairwise kernel interactions between 
$x^{(i)}$ and the current subset points. When a point $x^{(j)} \in P_{\text{temp}}$ is removed, all $B_i$ values can be 
updated efficiently by subtracting the terms involving $x^{(j)}$; conversely, when 
adding a new point, the same updates are performed in reverse by addition. Both updates take only $\Theta(dn)$ time.  

To identify the optimal swap pair $(x^{(h)}, x^{(k)})$, with 
$x^{(h)} \in P_{\text{temp}}$ and $x^{(k)} \notin P_{\text{temp}}$, we evaluate
\[
d(P_{\text{temp}}) + B_k - B_h + g(h,k),
\]
where $g(h,k)$ captures the interaction between the two points and can be computed 
in $\Theta(d)$ time as $g(h,k) = -V(h,k) - V(k,h)$. 
Since there are $(n - m)m$ such candidate pairs, one full swap iteration costs 
$\Theta(d(n - m)m)$, or simply $\Theta(dnm)$ when $n \gg m$. For smaller populations, all $g(h,k)$ values can be precomputed and stored in an 
$n \times n$ matrix, further reducing each swap step to $\Theta(nm)$ time. 
However, in large-scale settings (e.g., $n \ge 10^6$), this approach becomes 
impractical due to memory constraints.

\paragraph{Restarts and Initialization}
This swap-based mechanism forms the core of our subset optimization algorithm. 
On top of this, we introduce two parameters to control the number of 
\emph{local} and \emph{global} restarts, denoted by $R_\ell$ and $R_g$, respectively, 
as well as the number the \emph{initial populations} $L$. 
Global restarts correspond to entirely new optimization runs, each beginning from 
a different randomly selected subset of size $m$. 
Local restarts, in contrast, act as perturbations within a single global run: once 
a local optimum is reached (i.e., no improving swap exists), we randomly replace a 
small number of points—typically fewer than 10 even for large $m$—in the current 
best subset and resume the swap-based optimization from this perturbed configuration. 
The number of both global and local restarts are user-defined and allow for  trade-offs between runtime and robustness against poor local minima. To further diversify the initialization stage, we generate several ($L \approx 40$) 
random subsets and select the one that is most dissimilar from previously used 
initializations while achieving the smallest objective value $d(P_{\text{temp}})$. 

Overall, this combination of swap-based local search, fast updates, and controlled restarts provides an effective and scalable heuristic for subset selection under general kernel discrepancy objectives.

\section{Results}

We now assess the performance of the subset selection heuristic as presented in Section \ref{sec:algo}. Specifically, we optimize three kernel discrepancy objective functions; the regular $L_2$ star discrepancy, the generalized $L_2$ star discrepancy and the KSD. We consider a range of instances of $m$ from small element sets to a few hundred and in varying dimension, commonly settling for results in two dimensions. 

In the classical setting when obtaining uniform low-discrepancy point sets, we compare the star discrepancy values of the generated point sets to those of the $L_{\infty}$ subset selection results of~\cite{CDP22,CDP23}, Message-Passing Monte Carlo (MPMC)~\cite{RUSCHKIRK2024}, and traditional constructions such as the lifted Kronecker set for golden ratio and the Sobol' sequence. When targeting nonuniform distributions $F$, we compare the KSD of the subset selected sets against the already established method of \textit{Stein Points} \cite{STEINPOINTS}; see \ref{app:Steinpoints}.

\subsection{Experimental Implementation}

\paragraph{Initialization Sets} The results for the $L_2$ star discrepancies use a prefix of the Sobol' sequence as the initial population set. In the results presented, we select from the first 10 million points of the sequence. For non-uniform target distributions, standard low-discrepancy constructions are generally not available. In such cases, we typically rely on an IID random sample for initialization; otherwise, samples generated via MCMC are also acceptable (see Section \ref{sec:mcmc_appl}).

\paragraph{$L_2$ Star Discrepancy} For experiments based on the classical $L_2$ star discrepancy, we employ both the 
original star discrepancy kernel and its weighted extension as introduced by Hickernell \cite{hickernell1998generalized}. 
The unweighted case corresponds to the standard product kernel
\[
k(x, x') = \prod_{j=1}^d \big( 1 - \max(x_j, x'_j) \big),
\]
and the weighted star discrepancy kernel
\[
k(x, x') = \prod_{j=1}^d 1+\gamma_j\left( 1 - \max(x_j, x'_j) \right),
\]
where $\gamma_j$ for $1\leq j \leq d$ are the product weights assigned to each coordinate. In all reported experiments, we take $\gamma_j = 1$ since we do not have any specific apriori integrand knowledge to exploit, thus ensuring that all coordinates contribute equally to the discrepancy.

\paragraph{Kernel Stein Discrepancy} When working with the kernel Stein discrepancy, an important choice is the base kernel $k$. To avoid confounding results, we use the standard Radial Basis Function (RBF) kernel
\[
k(x,x') = \exp \left( - \frac{\|x-x'\|^2}{2h^2} \right)
\]
as the base kernel function in \eqref{eq:KSD} where the bandwidth parameter $h$ is chosen via the median heuristic across all experiments. Specifically, we take the bandwidth to be $h = \sqrt{\text{med}^2/2\log (N+1)}$ where $\text{med}$ is the median of the pairwise distances between the current samples of the point set. This choice is motivated to ensure the fairest comparison possible across methods.

\subsection{Experimental Results}\label{sec:expe}

\subsubsection{Optimizing the Star Discrepancy}
All experiments presented in this section were run with a one hour runtime budget on individual cores of the MeSU cluster of Sorbonne Université. Each run could choose its starting subset from 100 randomly sampled subsets, and had a budget of 5 general restarts and 5 local restarts. The final set returned is the one with the lowest $L_2$ star discrepancy found. All $L_{\infty}$ star discrepancies are then evaluated exactly with the DEM algorithm~\cite{DEM}.

\begin{figure}[t]
    \centering
    \includegraphics[width=0.5\linewidth]{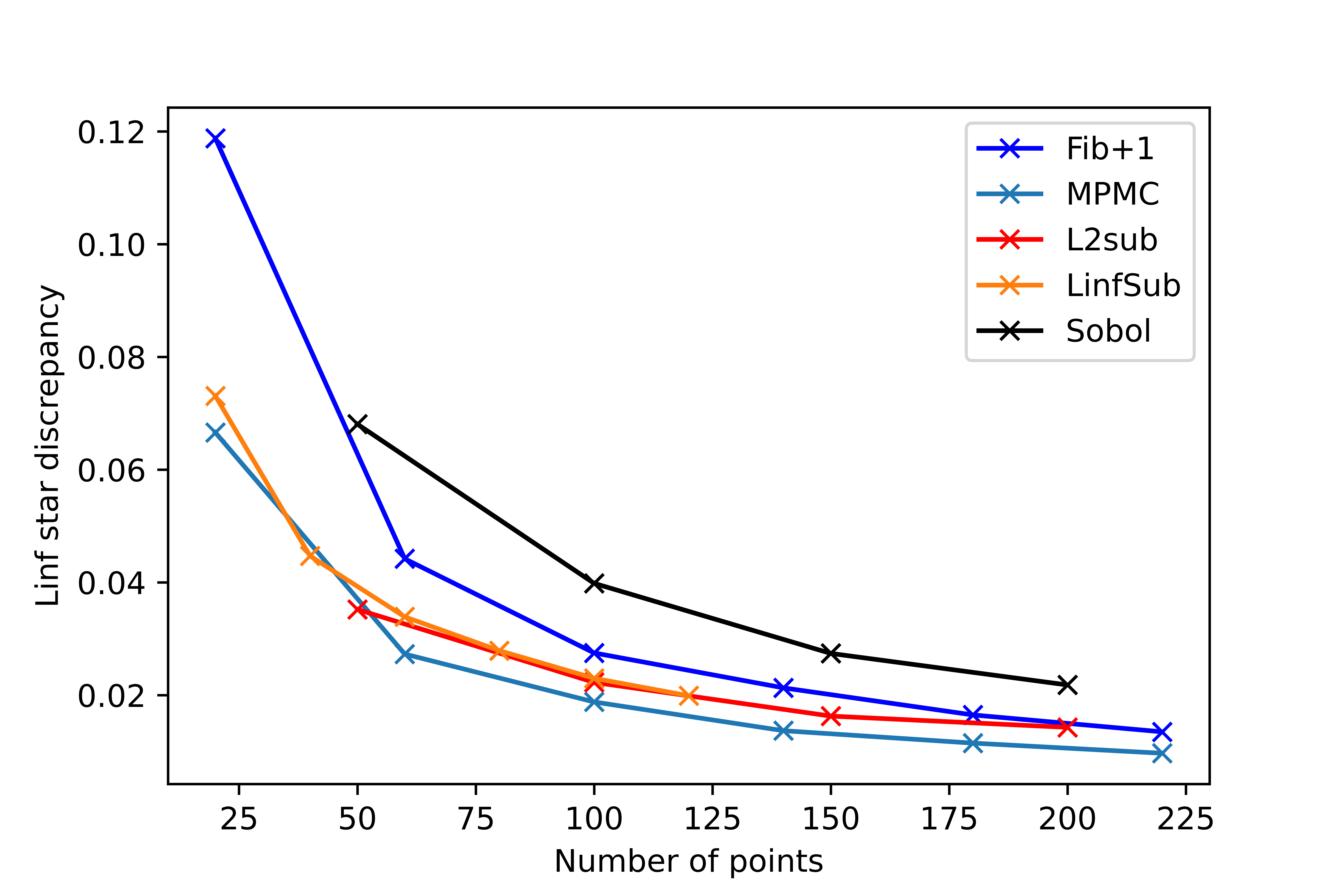}
    \caption{$L_{\infty}$ star discrepancy comparison in two dimensions of the different construction and optimization methods.}
    \label{fig:2d}
\end{figure}

Figure~\ref{fig:2d} compares the results of subset selection for the $L_2$ star discrepancy to recent approaches such as $L_{\infty}$ subset selection~\cite{CDP22,CDP23} and~\cite{RUSCHKIRK2024}, as well as with the discrepancy of the Sobol' sequence and the Fibonacci set: $\left\{(k/m,\{k\varphi\}):k \in \left\{0,\ldots,m-1\right\}\right\}$. 

First, $L_2$ subset selection does not perform as well as MPMC (and a fortiori as well as~\cite{PnasCDKP}). However, it does match the performance of the exact version of $L_{\infty}$ subset selection, while we are optimizing purely the $L_2$ star discrepancy. It is also possible that a better set for the $L_{\infty}$ star discrepancy was found at some iteration of the algorithm, but discarded in favor of one with lower $L_2$ star discrepancy. While there is a noticeable improvement relative to the Sobol' sequence initially, this decreases as $m$ increases. Table~\ref{tab:2dapp} in \ref{app:extended2D} details some extended results in two dimensions and shows that $L_2$ subset even becomes \emph{worse} than Sobol' for specific higher values of $m$. This could be linked to the relatively small, fixed, number of restarts and the increasing number of local optima as the subset size increases, as well as the limited total runtime.

For dimensions 3 and higher, there are fewer benchmark results available for comparison. Tables~\ref{tab:3d} and~\ref{tab:4d} present results for the $L_2$ subset selection method alongside the $L_{\infty}$ subset selection and the initial Sobol' set of equal size. As in two dimensions, $L_2$ subset selection consistently improves upon the corresponding Sobol' prefix. Notably, and perhaps one of our most surprising results, it now \emph{outperforms} $L_{\infty}$ subset selection across all tested instances, often by a considerable margin. Moreover, unlike the $L_{\infty}$ approach, our method is not restricted to small sample sizes: $L_2$ subset selection readily produces low-discrepancy sets containing thousands of points. However, as in two dimensions, performance gains relative to Sobol' diminish with increasing set size—showing only marginal improvement for 1000 points in three dimensions. This limitation is likely algorithmic rather than fundamental and could be alleviated by increasing the allowed runtime or number of restarts.

\begin{table}[h]
    \centering
    \begin{tabular}{|c|c|c|c|}
    \hline
       Number of points  & $L_2$ Subset & $L_{\infty}$ Subset & Sobol' \\
       \hline
        50 & 0.05952 & 0.080156 & 0.09708\\
        
        100 & 0.03835 & 0.048574 & 0.06058\\
        150 & 0.02612 & 0.035248 & 0.04483\\
        200 & 0.02203 & 0.028038 & 0.03315\\
        250 & 0.01840 & -& 0.02548\\
        500 & 0.01207 & -& 0.01460\\
        1000 & 0.00810&-& 0.008640\\
        2000 & 0.005030 &- & 0.005247 \\
        \hline
    \end{tabular}
    \caption{$L_{\infty}$ star discrepancy values obtained via $L_2$ and $L_{\infty}$ subset selection methods, compared to the Sobol' set of the same size in 3 dimensions. The $L_2$ subset always gives the lowest $L_{\infty}$ star discrepancy value.}
    \label{tab:3d}
\end{table}

\begin{table}[h]
    \centering
    \begin{tabular}{|c|c|c|c|}
    \hline
       Number of points  & $L_2$ Subset & $L_{\infty}$ Subset & Sobol' \\
       \hline
        50 &0.08482 & 0.097189 & 0.13422\\
        100 & 0.04760& 0.061478 & 0.09269\\
        150 & 0.04110& 0.053195 & 0.06174\\
        200 & 0.03008& 0.043015 & 0.05026\\
        250 & 0.02596& - & 0.03822\\
        500 & 0.01810& -& 0.02290\\
        1000 & 0.011727 & -& 0.014759\\
        2000 & 0.007808 &-& 0.008398\\
        \hline
    \end{tabular}
    \caption{$L_{\infty}$ star discrepancy values obtained via $L_2$ and $L_{\infty}$ subset selection methods, compared to the Sobol' set of the same size in 4 dimensions. The $L_2$ subset always gives the lowest $L_{\infty}$ star discrepancy value.}
    \label{tab:4d}
\end{table}

In five dimensions, we additionally report results for the weighted $L_2$ star discrepancy subset selection. Somewhat unexpectedly, this weighted variant performs \emph{worse} than the standard (unweighted) $L_2$ discrepancy. This indicates that introducing coordinate weights—intended to emphasize important interactions customized for a specific problem—does not universally improve point set quality. While the weighted discrepancy can act as a useful $L_{\infty}$ surrogate in higher dimensions when the integrand structure is known (for instance, in the Asian option pricing experiments of~\cite{RUSCHKIRK2024}), its utility for generating domain filling designs is limited.

\begin{table}[h]
    \centering
    \begin{tabular}{|c|c|c|c|c|}
    \hline
    Number of points  & $L_2$ Subset& Weighted $L_2$ & $L_{\infty}$ Subset & Sobol' \\
    \hline
         50 & \textbf{0.115507}  & 0.138205 & 0.118428 & 0.165488\\
        100 & 0.070071 & 0.084502 &  \textbf{0.068499}& 0.120707\\
        150 & \textbf{0.055612} & 0.067193 &  0.064438& 0.074899\\
        200 & \textbf{0.043016}  & 0.058678 &  0.052454& 0.058292\\
        250 & \textbf{0.034895}  & 0.047364 & -&0.053507 \\
        500 & \textbf{0.026379} & 0.031544&- &0.029017 \\ 
        1000 & 0.018813 & 0.021454& - & \textbf{0.018411}\\
        \hline
    \end{tabular}
    \caption{$L_{\infty}$ star discrepancy values in 5 dimensions for, from left to right, $L_2$ subset selection, weighted $L_2$ subset selection, $L_{\infty}$ subset selection and the Sobol' sequence. For each $n$, the best value is in bold.}
    \label{tab:5d}
\end{table}

\subsubsection{Optimizing the KSD}

Moving to optimize the KSD, we focus our results on two-dimensional nonuniform distributions where the density functions are available analytically; see \ref{app:distributions} for full details of the target distributions. We select from smaller population sizes, $n=1000$ and $n=10000$, than in the classical $L_2$ star discrepancy experiments and aim for a few hundred low-discrepancy points. Opting for smaller population sizes in this experiment enables us to evaluate how sensitive our subset selection algorithm is to having ``good choices" during initialization.

\begin{figure}[t]
  \centering
  \begin{subfigure}[b]{0.33\textwidth}
    \includegraphics[width=\textwidth]{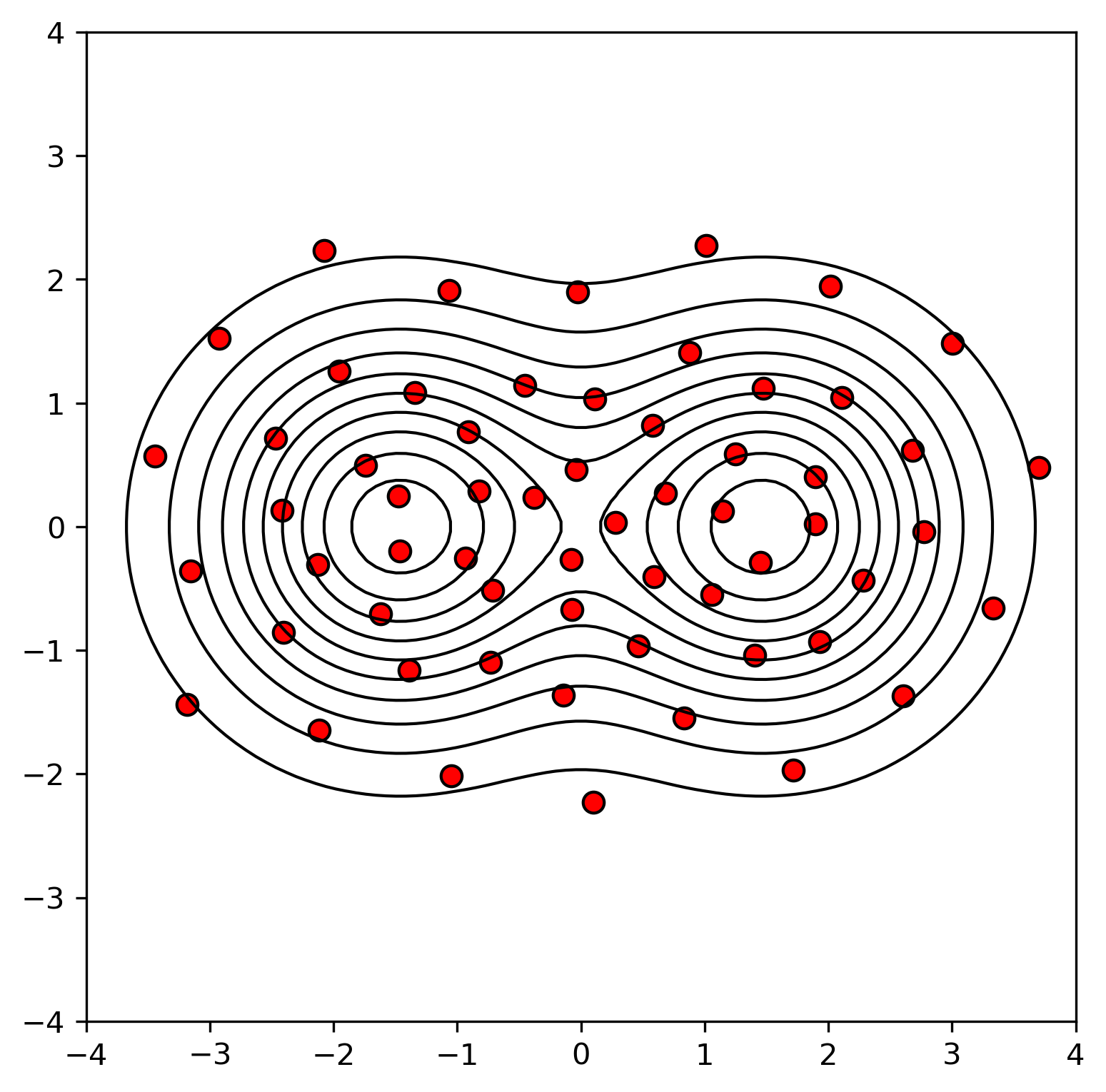}
    \label{fig:gauss}
  \end{subfigure}
  \hspace{1cm}
  \begin{subfigure}[b]{0.33\textwidth}
    \includegraphics[width=\textwidth]{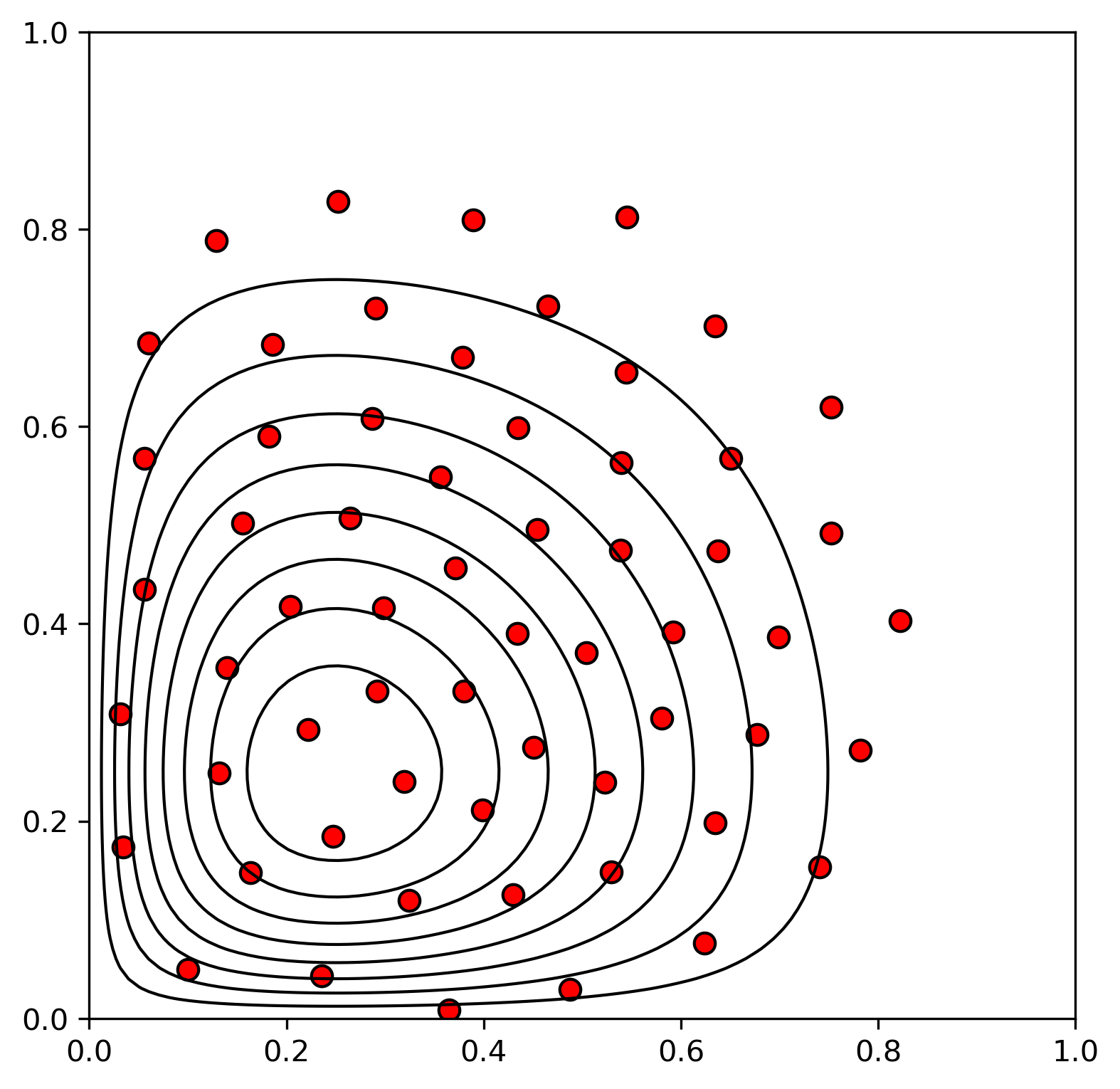}
    \label{fig:beta}
  \end{subfigure}
  \caption{Subset Selected point sets with respect to KSD for $n=60$. (\textbf{Left}) Mixture of two Gaussians  (\textbf{Right}) Product of independent Beta distributions.} 
  \label{fig:KSD_points}
\end{figure}

Figure \ref{fig:KSD_points} demonstrates the effectiveness of our method when applied to samples from a Beta-product distribution and a Gaussian mixture distributions;  Compared against randomly selected subsets of size $m = 20,60,100,\ldots,500$ from larger sample pools ($n = 1000$ and $n = 10000$), and a baseline using Stein Points, our method consistently achieves lower squared KSD. The reduction is particularly pronounced for small $m$, where selection has the most significant impact on quality. Notably, for the Beta distribution example, our method achieves 50\% improvement in squared KSD compared to Stein Points at $m = 500$ when selecting from the larger pool of samples. Our results also confirm the intuitive conclusion that selecting from a larger candidate set improves the performance of our selection algorithm, as seen from the consistently lower KSD values for $n = 10000$ compared to $n = 1000$. The generated subsets (Figure \ref{fig:KSD_points}) align well with the target densities in a visual assessment.

We note in passing that the KSD is known to exhibit certain undesirable pathologies. In particular, it can admit spurious minima corresponding to configurations where points cluster around saddle regions of the target density rather than its true modes. This issue is detailed elegantly by \cite{KSDpathologies,KSDdescent}, who propose a remedy based on augmenting the KSD objective with a second-order (Laplacian) regularization term to suppress such degenerate solutions. In principle, this modification could be incorporated within our framework, especially for highly multi-modal target distributions where saddle-point equilibria are more likely to occur.

\begin{table}[h]
    \centering
    \begin{tabular}{|c|c|c|c|}
    \hline
    Number of points  & Subset ($n=1000$) & Subset ($n=10000$) & Stein Points  \\
    \hline
         10 & 0.31391 & 0.35244 & 0.338273  \\
        25 & 0.15716 & 0.15008 & 0.23851  \\
        50 & 0.12430 & 0.10749 & 0.21139 \\
        75 & 0.11306 & 0.09576 & 0.18728  \\
        100 & 0.11060 & 0.09006 & 0.17630  \\
        125 & 0.10516 & 0.08718 & 0.17592 \\ 
        150 & 0.10586 & 0.08401 & 0.16202 \\
        175 & 0.10809 & 0.08167 & 0.15339 \\
        200 & 0.10661 & 0.08040 & 0.15103 \\
        225 & 0.10543 & 0.07942 & 0.14976 \\
        \hline
    \end{tabular}
    \caption{(Squared) KSD for mixture of Gaussian problem for two subset selection procedures with different population sizes and the Stein Points method.}
    \label{tab:stein_disc}
\end{table}

\section{Discussion}\label{sec:discussion}

Encouraged by recent work exploring $L_2$-based discrepancy measures \cite{clément2025optimizationdiscrepancymeasures, RUSCHKIRK2024}, we present a subset selection algorithm that leverages the structure of general kernel discrepancies, such as the KSD and $L_2$ star discrepancy, to select representative, low-discrepancy point sets from large set initializations. By moving away from the previously explored $L_\infty$-based discrepancy measures, our approach benefits from smooth and decomposable objectives that enable efficient optimization and adaptation to non-uniform target distributions via a RHKS framework. In the following subsections, we now make several observations and comments on our work, its applicability and future directions.

\subsection{Correlation between the $L_2$ and $L_{\infty}$ star discrepancy}\label{sec:L2forLinf}

Across all experiments, the $L_2$ subset selection method generally achieves lower $L_{\infty}$ star discrepancy than the $L_{\infty}$-optimized counterpart—an unexpected outcome given that performance is measured under the $L_{\infty}$ criterion itself. The only exception occurs in the five-dimensional 100-point case, where $L_{\infty}$ subset selection slightly outperforms $L_2$ selection. This empirical trend suggests a deep alignment between the $L_2$ and $L_{\infty}$ notions of uniformity: optimizing the simpler $L_2$ objective often suffices to produce excellent configurations for $L_{\infty}$ as well. The results in this paper are also consolidated by separate studies; the MPMC framework \cite{RUSCHKIRK2024} also relies on $L_2$-based optimization and yet attains low $L_{\infty}$ discrepancy, while Kritzinger’s greedy one-dimensional construction~\cite{CKritz,Kritz} exhibits the same effect.

These findings motivate two directions for further investigation.
First, a sharp theoretical connection between the $L_2$ and $L_{\infty}$ star discrepancies remains to be precisely quantified. Second, the search for a better surrogate in high dimensions is entirely open. Although $L_2$ subset selection continues to perform acceptably up to dimension five with respect to the $L_\infty$ star discrepancy, its advantage diminishes thereafter. Weighted extensions such as those proposed in~\cite{RUSCHKIRK2024} appear ineffective as general-purpose substitutes for $L_{\infty}$ except when the coordinate weights are chosen judiciously to target a specific class of problems. At a broader level, even the centrality of the $L_{\infty}$ star discrepancy itself warrants further re-examination as is highlighted in \cite{clément2025optimizationdiscrepancymeasures}. In high dimensions, anchored test boxes occupy vanishingly small volume, calling into question whether the $L_{\infty}$ measure remains the most meaningful indicator of uniformity in practice.

\subsection{Problem reformulation for exact approaches}

This paper focused entirely on non-exact, heuristic methods for subset selection. We finish by mentioning several possible methods for solving the optimization problem \eqref{eq:optim_problem} exactly which act as candidates for future work. Further details can be found in~\cite{PhDFC}.

As mentioned several times already, the contribution of each point to the total discrepancy can be decomposed into two components, a self-contribution \( V(i,i) \), and a set of pairwise interaction terms \( V(i,j) \) for \( i \ne j \) as in \eqref{eq:contribution}.
The first proposed exact method approaches the subset problem as a graph problem. Viewing each point as a node, the self-contributions \( V(i,i) \) become node weights, while the pairwise terms \( V(i,j) \) for \( i\ne j\) define edge weights in a complete weighted graph. Subset selection then becomes equivalent to finding a \emph{minimum-weight \( m \)-clique} \cite{cliqueproblem}, with a \emph{very} specific set of weights.

The second approach is to re-formulate as an instance of the \emph{Unconstrained Binary Quadratic Problem (UBQP)}~\cite{Kochen}. While subset selection imposes a cardinality constraint, this can be incorporated into the diagonal weights.

\subsection{Complexity}
Both approaches proposed in the previous section raise a natural complexity-theoretic question: is the subset selection problem for kernel discrepancies NP-hard? It is well known that related problems (such as subset selection for the \( L_\infty \) star discrepancy) are NP-hard~\cite{CDP22}. Separately, clique problems are also traditionally hard~\cite{Karp1972}. In contrast, any kernel discrepancy of a given point set can be computed in polynomial time, and once the weight matrix has been constructed (which takes \( O(dn^2) \) time), the dimension \( d \) no longer affects the complexity of subset selection. However, due to the highly structured and correlated nature of the entries in weight matrix, the complexity of kernel discrepancy-based subset selection problem remains unresolved.

\begin{conjecture}
    Let $m,n\in \mathbb{N}$ with $m < n$ and $d$ be a general kernel discrepancy as in \eqref{eq:general_disc}. The subset selection problem 
    \[
    \argmin_{P_m \subseteq P,\ |P_m| = m} d(P_m)
    \]
    is NP-hard.
\end{conjecture}

\subsection{Application to MCMC Thinning}\label{sec:mcmc_appl}

In Markov Chain Monte Carlo (MCMC) methods, one typically draws a long sequence of samples from a target distribution which may be highly correlated, especially when the chain mixes slowly. To reduce the effect of this correlation, \emph{thinning} is sometimes performed which refers to selecting a subset of the samples with the goal of obtaining a representative sample that behaves more like an IID sample from the target distribution, or has better statistical properties. Thinning is often controversial when done naively (e.g., taking every 10th sample) because it discards data blindly.

Although our method is not designed specifically as an MCMC thinning algorithm and sits outside of the scope of this work, it could naturally be used as one in settings where a sample has been drawn from a distribution, such as via MCMC, and the target density function is known or can be evaluated. Rather than relying on heuristic thinning strategies (e.g., taking every $k^{th}$ point), our approach would select a subset by directly optimizing the KSD. Related work exists in the literature as \cite{KSDpathologies,steinthinning2022} employ different optimization strategies for Stein discrepancies.

\paragraph{Acknowledgments} 
Computations were performed on the SACADO MeSU platform at Sorbonne Université. FC and CD are grateful to Noam Rosen for initial experiments that motivated this work. 
The work of CD is financially supported by the European Union (ERC, ``dynaBBO'', grant no.~101125586). Views and opinions expressed are however those of the authors only and do not necessarily reflect those of the European Union or the European Research Council Executive Agency. Neither the European Union nor the granting authority can be held responsible for them. 
The work of NK was supported by the National Science Foundation (DMS Grant No. 2316011).

\bibliographystyle{alpha}
\bibliography{refs,refs_NK}

\newcommand{\etalchar}[1]{$^{#1}$}
\begin{thebibliography}{KAFMA21}

\bibitem[AM23]{GradientFreeKSD2023}
E.~Afzali and S.~Muthukumarana.
\newblock Gradient-free kernel conditional {S}tein discrepancy goodness of fit testing.
\newblock {\em Machine Learning with Applications}, 12:100463, 2023.

\bibitem[BBD{\etalchar{+}}19]{MINSteinDisc2019}
A.~Barp, F.-X. Briol, A.~Duncan, M.~Girolami, and L.~Mackey.
\newblock Minimum {S}tein discrepancy estimators.
\newblock In H.~Wallach, H.~Larochelle, A.~Beygelzimer, F.~d\textquotesingle Alch\'{e}-Buc, E.~Fox, and R.~Garnett, editors, {\em Advances in Neural Information Processing Systems}, volume~32. Curran Associates, Inc., 2019.

\bibitem[BK73]{cliqueproblem}
C.~Bron and J.~Kerbosch.
\newblock Algorithm 457: finding all cliques of an undirected graph.
\newblock {\em Commun. ACM}, 16(9):575–577, 1973.

\bibitem[BLV08]{BilykSmall}
D.~Bilyk, M.T. Lacey, and A.~Vagharshakyan.
\newblock On the small ball inequality in all dimensions.
\newblock {\em J. Funct. Anal.}, 254:2470--2502, 2008.

\bibitem[BSDV23]{KSDpathologies}
C.~B\'{e}nard, B.~Staber, and S.~Da~Veiga.
\newblock Kernel stein discrepancy thinning: a theoretical perspective of pathologies and a practical fix with regularization.
\newblock In {\em Proceedings of the 37th International Conference on Neural Information Processing Systems}, NIPS '23, Red Hook, NY, USA, 2023. Curran Associates Inc.

\bibitem[CCD{\etalchar{+}}20]{CauwetCDLRRTTU20}
M.{-}L. Cauwet, C.~Couprie, J.~Dehos, P.~Luc, J.~Rapin, M.~Rivi{\`{e}}re, F.~Teytaud, O.~Teytaud, and N.~Usunier.
\newblock Fully parallel hyperparameter search: Reshaped space-filling.
\newblock In {\em Proc. of the 37th {ICML}}, volume 119 of {\em PMLR}, pages 1338--1348. {PMLR}, 2020.

\bibitem[CDKP25]{PnasCDKP}
F.~Clément, C.~Doerr, K.~Klamroth, and L.~Paquete.
\newblock Searching permutations for constructing uniformly distributed point sets.
\newblock {\em Proceedings of the National Academy of Sciences}, 122(14):e2424464122, 2025.

\bibitem[CDP22]{CDP22}
F.~Cl\'{e}ment, C.~Doerr, and L.~Paquete.
\newblock Star discrepancy subset selection: {P}roblem formulation and efficient approaches for low dimensions.
\newblock {\em Journal of Complexity}, 70:101645, 2022.

\bibitem[CDP24]{CDP23}
F.~Cl\'{e}ment, C.~Doerr, and L.~Paquete.
\newblock Heuristic approaches to obtain low-discrepancy point sets via subset selection.
\newblock {\em Journal of Complexity}, 83:101852, 2024.

\bibitem[CKOR25]{clément2025optimizationdiscrepancymeasures}
F.~Clément, N.~Kirk, A.~B. Owen, and T.~K. Rusch.
\newblock On the optimization of discrepancy measures.
\newblock {\em Preprint, arXiv 2508.04926}, 2025.
\newblock https://arxiv.org/abs/2508.04926.

\bibitem[CKP24]{KIRKPAUSINGERCLEMENT2024}
F.~Clément, N.~Kirk, and F.~Pausinger.
\newblock Partitions for stratified sampling.
\newblock {\em Monte Carlo Methods and Applications}, 30(2):163--181, 2024.

\bibitem[Cl{\'{e}}23]{CKritz}
F.~Cl{\'{e}}ment.
\newblock Outperforming the best {1D} low-discrepancy constructions with a greedy algorithm.
\newblock 2023.
\newblock https://arxiv.org/abs/2406.18132.

\bibitem[Cl{\'{e}}24]{PhDFC}
F.~Cl{\'{e}}ment.
\newblock {\em An Optimization Perspective on the Construction of Low-Discrepancy Point Sets}.
\newblock PhD thesis, Sorbonne Université, 2024.

\bibitem[CMG{\etalchar{+}}18]{STEINPOINTS}
W.~Y. Chen, L.~Mackey, J.~Gorham, F.-X. Briol, and C.~Oates.
\newblock Stein points.
\newblock In Jennifer Dy and Andreas Krause, editors, {\em Proceedings of the 35th International Conference on Machine Learning}, volume~80 of {\em Proceedings of Machine Learning Research}, pages 844--853. PMLR, 10--15 Jul 2018.

\bibitem[CRPR25]{chahine2025improvingefficiencysamplingbasedmotion}
M.~Chahine, T.~K. Rusch, Z.~J. Patterson, and D.~Rus.
\newblock Improving {E}fficiency of {S}ampling-based motion planning via {M}essage-{P}assing {M}onte {C}arlo.
\newblock In {\em 9th Annual Conference on Robot Learning}, 2025.

\bibitem[CST14]{Pano}
W.~Chen, A.~Srivastav, and G.~Travaglini.
\newblock {\em A Panorama of Discrepancy Theory}, volume 2107.
\newblock 01 2014.

\bibitem[DEM96]{DEM}
D.~P. Dobkin, D.~Eppstein, and D.~P. Mitchell.
\newblock Computing the discrepancy with applications to supersampling patterns.
\newblock {\em ACM Trans. Graph.}, 15(4):354–376, oct 1996.

\bibitem[dNVBK24]{DiedOpti}
J.~de~Nobel, D.~Vermetten, T.~B{\"a}ck, and A.~Kononova.
\newblock Sampling in {CMA-ES}: Low numbers of low discrepancy points.
\newblock {\em Proc. of IJCCI 2024}, pages 120--128, 2024.

\bibitem[DP10]{DickP10}
J.~Dick and F.~Pillichshammer.
\newblock {\em Digital Nets and Sequences}.
\newblock Cambridge University Press, Cambridge, 2010.

\bibitem[FH14]{FangHick2014}
K.-T. Fang and F.~J. Hickernell.
\newblock {\em Uniform Experimental Designs}.
\newblock John Wiley \& Sons, Ltd, 2014.

\bibitem[GJ97]{GalFin}
S.~Galanti and A.~Jung.
\newblock Low-discrepancy sequences: {Monte-Carlo} simulation of option prices.
\newblock {\em J. Deriv}, pages 63--83, 1997.

\bibitem[GKWW12]{W1hard}
P.~Giannopoulos, C.~Knauer, M.~Wahlstr{\"{o}}m, and D.~Werner.
\newblock Hardness of discrepancy computation and $\epsilon$-net verification in high dimension.
\newblock {\em Journal of Complexity}, 28:162--176, 2012.

\bibitem[GM15]{GorhamMackey2015}
J.~Gorham and L.~Mackey.
\newblock Measuring sample quality with {S}tein’s method.
\newblock In {\em Cortes, C., Lawrence, N. D., Lee, D. D., Sugiyama, M., and Garnett, R. (eds.), Adv. NIPS 28}, pages 226--234. Curran Associates, Inc., 2015.

\bibitem[GSW09]{complexity}
M.~Gnewuch, A.~Srivastav, and C.~Winzen.
\newblock Finding optimal volume subintervals with k points and calculating the star discrepancy are {NP}-hard problems.
\newblock {\em Journal of Complexity}, 25(2):115--127, 2009.

\bibitem[GWW12]{GnewuchWW12}
M.~Gnewuch, M.~Wahlstr{\"{o}}m, and C.~Winzen.
\newblock A new randomized algorithm to approximate the star discrepancy based on threshold accepting.
\newblock {\em {SIAM} J. Numerical Analysis}, 50(2):781--807, 2012.

\bibitem[Hal60]{Halton60}
J.H. Halton.
\newblock On the efficiency of certain quasi-random sequences of points in evaluating multi-dimensional integrals.
\newblock {\em Numerische Mathematik}, 2:84--90, 1960.

\bibitem[Hic98]{hickernell1998generalized}
F.~Hickernell.
\newblock A generalized discrepancy and quadrature error bound.
\newblock {\em Mathematics of computation}, 67(221):299--322, 1998.

\bibitem[HKC{\etalchar{+}}25]{vanhuffel2025neurallowdiscrepancysequences}
M.~E.~Van Huffel, N.~Kirk, M.~Chahine, D.~Rus, and T.~K. Rusch.
\newblock Neural low-discrepancy sequences.
\newblock {\em arXiv, Preprint}, 2025.
\newblock https://arxiv.org/abs/2510.03745.

\bibitem[Hla61]{Hlawka}
E.~Hlawka.
\newblock Funktionen von beschränkter {Variation} in der {Theorie} der {Gleichverteilung}.
\newblock {\em Ann. Mat. Pum Appl.}, 54:325--333, 1961.

\bibitem[KAFMA21]{KSDdescent}
A.~Korba, P.-C. Aubin-Frankowski, S.~Majewski, and P.~Ablin.
\newblock Kernel {S}tein discrepancy descent.
\newblock In Marina Meila and Tong Zhang, editors, {\em Proceedings of the 38th International Conference on Machine Learning}, volume 139 of {\em Proceedings of Machine Learning Research}, pages 5719--5730. PMLR, 18--24 Jul 2021.

\bibitem[Kar72]{Karp1972}
R.~M. Karp.
\newblock {\em Reducibility among Combinatorial Problems}, pages 85--103.
\newblock Springer US, Boston, MA, 1972.

\bibitem[KHG{\etalchar{+}}14]{Kochen}
G.~Kochenberger, J.{-}K. Hao, F.~Glover, M.~Lewis, Z.~L\"{u}, H.~Wang, and Y.~Wang.
\newblock The unconstrained binary quadratic programming problem: A survey.
\newblock {\em Journal of Combinatorial Optimization}, 28, 07 2014.

\bibitem[KN74]{KUIPNIED1974}
L.~Kuipers and H.~Niederreiter.
\newblock {\em Uniform distribution of sequences}.
\newblock Pure and Applied Mathematics. Wiley-Interscience [John Wiley \& Sons], New York-London-Sydney, 1974.

\bibitem[Kok43]{Koksma}
J.F. Koksma.
\newblock A general theorem from the theory of the uniform distribution modulo 1.
\newblock {\em Mathematica B (Zutphen)}, 1:7--11, 1942/1943.

\bibitem[KP23]{KIRKPAUSINGER2023}
N.~Kirk and F.~Pausinger.
\newblock On the expected ${L}_2$-discrepancy of jittered sampling.
\newblock {\em Unif. Dist. Theory}, 18(1):65--82, 2023.

\bibitem[Kri22]{Kritz}
R.~Kritzinger.
\newblock Uniformly distributed sequences generated by a greedy minimization of the ${L}_2$ discrepancy.
\newblock {\em Moscow Journal of Combinatorics and Number Theory}, 11(3):215--236, 2022.
\newblock https://arxiv.org/abs/2109.06298.

\bibitem[LLJ16]{LIULEEJORDAN2016}
Q.~Liu, J.~Lee, and M.~Jordan.
\newblock A kernelized stein discrepancy for goodness-of-fit tests.
\newblock In {\em International Conference on Machine Learning (ICML)}, pages 276--284, 2016.

\bibitem[LW16]{SVGD2016}
Q.~Liu and D.~Wang.
\newblock Stein variational gradient descent: A general purpose bayesian inference algorithm.
\newblock In {\em NeurIPS}, 2016.

\bibitem[Mat98]{MatL2}
J.~Matou{\v{s}}ek.
\newblock On the {L2}-discrepancy for anchored boxes.
\newblock {\em J. Complexity}, 14:527--556, 1998.

\bibitem[Mat10]{Mat}
J.~Matou{\v{s}}ek.
\newblock Geometric discrepancy.
\newblock {\em 2nd edition, Springer Berlin}, 2010.

\bibitem[Nie92]{Nie92}
H.~Niederreiter.
\newblock {\em Random Number Generation and Quasi-{M}onte {C}arlo Methods}, volume~63 of {\em SIAM CBMS-NSF Regional Conference Series in Applied Mathematics}.
\newblock SIAM, Philadelphia, 1992.

\bibitem[PBC{\etalchar{+}}22]{MatBuilder}
L.~Paulin, N.~Bonneel, D.~Coeurjoly, J.-C. Iehl, A.~Keller, and V.~Ostromoukhov.
\newblock {MatBuilder}: Mastering sampling uniformiy over projections.
\newblock {\em ACM Transactions on Graphics (proceedings of SIGGRAPH)}, 2022.

\bibitem[RCC{\etalchar{+}}22]{steinthinning2022}
M.~Riabiz, W.~Y. Chen, J.~Cockayne, P.~Swietach, S.~A. Niederer, L.~Mackey, and C.~J. Oates.
\newblock Optimal thinning of {MCMC} output.
\newblock {\em Journal of the Royal Statistical Society Series B: Statistical Methodology}, 84(4):1059--1081, 2022.

\bibitem[RKB{\etalchar{+}}24]{RUSCHKIRK2024}
T.~K. Rusch, N.~Kirk, M.~Bronstein, C.~Lemieux, and D.~Rus.
\newblock Message-{P}assing {M}onte {C}arlo: {G}enerating low-discrepancy point sets via graph neural networks.
\newblock {\em Proceedings of the National Academy of Sciences}, 121(40):e2409913121, 2024.

\bibitem[Sob67]{Sobol}
I.M. Sobol.
\newblock {On the Distribution of Points in a Cube and the Approximate Evaluation of Integrals}.
\newblock {\em {USSR Computational Mathematics and Mathematical Physics}}, 7(4):86~--~112, 1967.

\bibitem[Ste72]{STEIN1972}
C.~Stein.
\newblock A bound for the error in the normal approximation to the distribution of a sum of dependent random variables.
\newblock In {\em Proceedings of the Sixth Berkeley Symposium on Mathematical Statistics and Probability, Volume 2: Probability Theory}, pages 583--602. University of California Press, 1972.

\bibitem[SWN03]{SantnerDoE}
T.~J. Santner, B.J. Williams, and W.~I. Notz.
\newblock {\em The {Design} and {Analysis} of {Computer} {Experiments}}.
\newblock Springer {Series} in {Statistics}. Springer, 2003.

\bibitem[vdC35]{vdC}
J.G. van~der Corput.
\newblock Verteilungsfunktionen {II}.
\newblock {\em Akad. Wetensch. Amsterdam Proc.}, 38:1058--1066, 1935.

\bibitem[War72]{Warnock}
T.T. Warnock.
\newblock Computational inverstigations of low-discrepancy point sets.
\newblock {\em in Applications of number theory to numerical analysis, ed. by S.K. Zaremba (Academic Press, New York)}, 1972.

\end{thebibliography}

\appendix

\section{Code}
The code used in this project is available at \url{https://github.com/de0ch/l2_subset}. Further instructions to run the code can be found in the README.

\section{Further Implementation Details and Numerical Results}
\subsection{Extended results in dimension 2}\label{app:extended2D}
We provide here precise numerical values associated with the results used for Figure~\ref{fig:2d}.
\begin{table}[h!]
    \centering
    \begin{tabular}{|c|c|c|c|}
    \hline
       Number of points  & $L_2$ Subset& Weighted $L_2$ subset & Sobol' \\
       \hline
        50 & \textbf{0.035219} & 0.048551 & 0.068086\\
        100& \textbf{0.022277} & 0.033885 & 0.039844\\
        150& \textbf{0.016292} & 0.025748 & 0.027410\\
        200& \textbf{0.014293} & 0.022075 & 0.021813\\
        250& \textbf{0.011852} & 0.016913 & 0.015680\\
        500& \textbf{0.006944} & 0.009694 & 0.008173\\
        1000 & 0.005040& 0.006689 & \textbf{0.004705}\\
        1500 & \textbf{0.004059} & 0.005084 & 0.004273\\
        2000 & 0.003519 & - & \textbf{0.002701}\\
        \hline
    \end{tabular}
    \caption{$L_{\infty}$ star discrepancy values in two dimensions obtained via $L_2$ star and weighted $L_2$ star subset selection. The Sobol' values are added as a reference, the best value for each $n$ is in bold. The value for $n=2000$ for weighted is missing as it did not manage to reach a single local optimum in one hour.}
    \label{tab:2dapp}
\end{table}

\subsection{Beta Distribution Results}

\begin{table}[h!]
    \centering
    \begin{tabular}{|c|c|c|c|}
    \hline
    Number of points & Subset ($n=1000$) & Subset ($n=10000$) & Stein Points  \\
    \hline
        10 & 16.3517 & 19.7382 & 18.0771  \\
        25 & 6.3977 & 6.6506 & 10.1144  \\
        50 & 4.5439 & 4.0138 & 9.1641 \\
        75 & 4.0879 & 3.5007 & 8.2064  \\
        100 & 4.0989 & 3.2833 & 7.1020  \\
        125 & 4.1848 & 3.2183 & 6.7280 \\ 
        150 & 4.0181 & 3.1468 & 6.5906 \\
        175 & 4.1326 & 3.1135 & 6.4906 \\
        200 & 4.0172 & 3.1403 & 6.4891 \\
        225 & 3.9418 & 3.0569 & 6.1045 \\
        \hline
    \end{tabular}
    \caption{(Squared) KSD for product of Beta distribution problem for subset selection procedures with different population sizes and the Stein Points method.}
    \label{tab:stein_disc_beta}
\end{table}

\subsection{Stein Points}\label{app:Steinpoints}

Computation of the $n^{th}$ Stein point is dependent upon the already existing $n-1$ terms and requires a global optimization to find $x^{(n)} \in \mathbb{R}^d$ that minimizes the kernel Stein discrepancy of the total $n$ element set, holding $(x^{(i)})_{i \in \{1,\ldots,n-1\}}$ fixed. In \cite{STEINPOINTS}, several numerical optimization methods are considered to solve this $\argmin$ problem. In our experiments, we implement the Adam optimizer with a learning rate of $0.01$, selected after testing several judiciously chosen alternatives for the learning rate.

\subsection{Nonuniform target distributions.}\label{app:distributions}
We first consider a Gaussian mixture model in two dimensions, which is somewhat of a standard benchmark for posterior inference methods. The target distribution is a two-component Gaussian mixture
\[
\frac{1}{2} \mathcal{N}(\mu_1, \Sigma_1) + \frac{1}{2} \mathcal{N}(\mu_2, \Sigma_2),
\]
where $\mu_1 = (-1.5,0), \mu_2 = (1.5,0), \Sigma_1 = \Sigma_2 = I$.

We also consider a Beta-distributed target density as an example of a bounded probability distribution. The target distribution is defined as the product of two independent Beta distributions
\[
X \sim \text{Beta}(\alpha_x, \beta_x), \quad Y \sim \text{Beta}(\alpha_y, \beta_y).
\]
This distribution is supported on \( (0,1) \times (0,1) \) and allows independent control over the shape of each marginal through the parameters \( \alpha_x, \beta_x, \alpha_y, \beta_y \). For our experiments, we set $\alpha_x = 2, \beta_x = 4, \alpha_y = 2, \beta_y = 4$.

\end{document}